\documentclass[10pt, a4paper]{article}
\usepackage[utf8]{inputenc}
\usepackage[T1]{fontenc}
\usepackage{geometry}
\usepackage{amsmath}
\usepackage{bm}
\usepackage{amssymb}
\usepackage{authblk}
\usepackage{hyperref}
\usepackage{caption}
\usepackage{algorithm}
\usepackage{algpseudocode}
\usepackage{graphicx}
\usepackage{tabularx}
\usepackage{subcaption}
\DeclareGraphicsExtensions{.pdf,.png,.jpg,.jpeg}
\graphicspath{{./Figures/}{./figures/}}
\usepackage{booktabs}  
\usepackage{tikz}
\usetikzlibrary{arrows.meta, positioning, fit}
\usepackage{setspace}
\usepackage[draft]{changes}
\usepackage[numbers]{natbib}
\usepackage{float}

\title{\textbf{Hybrid Neural Ordinary Differential Equations for Data-Efficient Polymerization Modeling with Incomplete Kinetics}}
\author[1]{Marah Almanasreh}
\author[2,1,3]{Alexander Mitsos}
\author[4,*]{Eike Cramer}

\affil[1]{\small RWTH Aachen University, Process Systems Engineering (AVT.SVT), 52074 Aachen, Germany}
\affil[2]{\small JARA-CSD, 52425 Jülich, Germany}
\affil[3]{\small Energy Systems Engineering (ICE-1), Forschungszentrum Jülich, 52425 Jülich, Germany}
\affil[4]{\small Department of Chemical Engineering,
Sargent Centre for Process Systems Engineering,
University College London,
Torrington Place, 
London WC1E 7JE, 
United Kingdom}
\affil[*]{Corresponding author: e.cramer@ucl.ac.uk}
\date{}

\begin{document}
\maketitle
\onehalfspacing
\begin{abstract}
\begin{figure}[H]
    \centering
    \includegraphics[width=\textwidth]{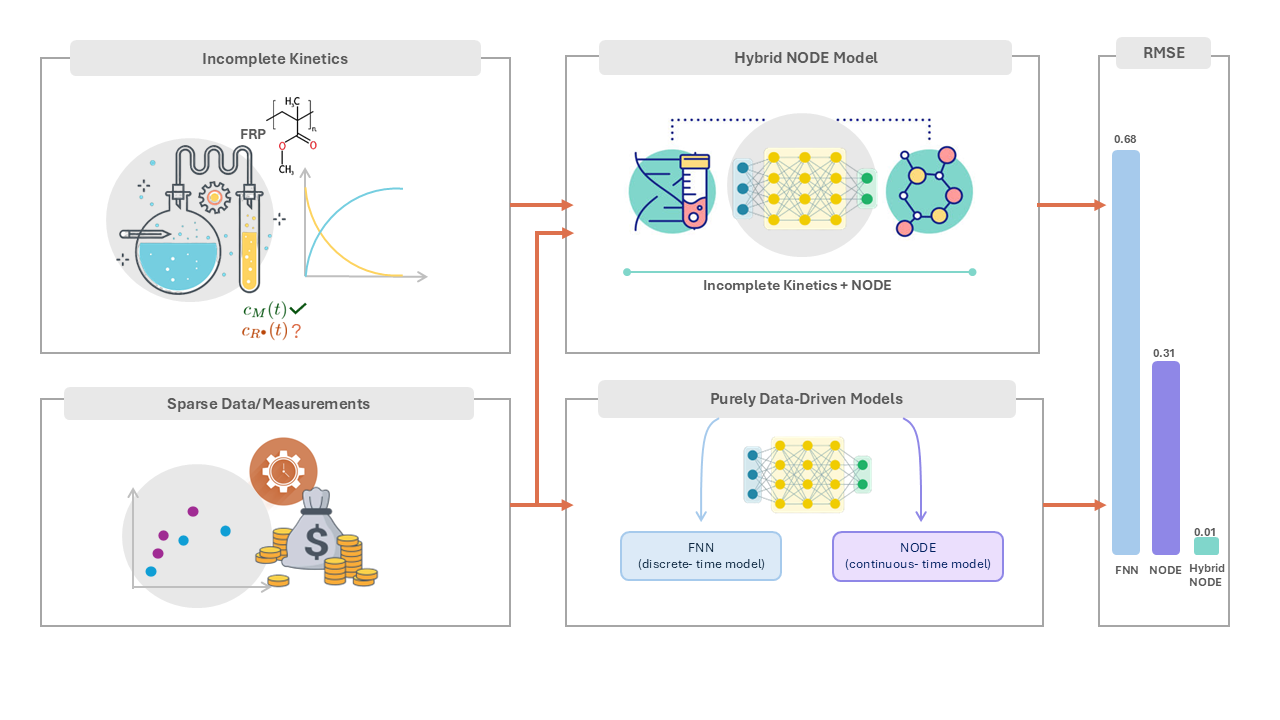}
    \label{fig:graphical_abstract}
\end{figure}
Accurate prediction of polymerization dynamics is essential for process design, control, and optimization. Yet, purely mechanistic models require labor-intensive parameterization of partially characterized kinetics, while purely data-driven models demand large, diverse datasets that are costly to obtain, particularly in early-design stages. We propose a hybrid Neural Ordinary Differential Equation (NODE) framework for data-efficient modeling of free-radical polymerization. Using batch polymerization of methyl methacrylate (MMA) as a case study, the mechanistic mass balances are retained explicitly, and only the partially-characterized effective radical concentration governing monomer consumption is learned from data through a neural network surrogate, while established reactions such as initiator decomposition, propagation, and termination remain physically modeled. The hybrid NODE is evaluated against a discrete-time feedforward neural network and a purely data-driven NODE under sparse data conditions, with models trained on as few as ten measurements under both regular and irregular sampling. The hybrid NODE consistently achieves lower prediction errors and more physically consistent extrapolations than both purely data-driven baselines. In a generalization scenario with noisy data and unseen operating conditions, the hybrid NODE achieves an RMSE of 0.013, compared to 0.31 for the data-driven NODE and 0.68 for the discrete-time model, demonstrating that learning only a closure term rather than the full dynamics is sufficient for reliable prediction under limited data availability.

\end{abstract}

\section{Introduction}
\label{Introduction}
Polymerization processes play a crucial role in producing a wide range of materials, from commodity plastics to high-performance polymers for advanced technologies. Accurate modeling and control of these processes are essential for optimizing production efficiency, enhancing product quality, and reducing manufacturing costs and environmental impacts~\cite{OHSHIMA2000135,karuppusamy2025review}. Among polymerization techniques, radical polymerization is particularly versatile and economically viable, which has led to its widespread industrial adoption~\cite{Asua2004, guerrero2025free}. 
However, the inherent complexity of radical polymerization kinetics involves numerous concurrent reaction pathways whose rates and radical population dynamics are sensitive to process conditions such as temperature and initiator concentration~\cite{Asua2004, Ballard2020}, posing significant challenges for accurate dynamic modeling. These sensitivities directly affect polymer quality, reactor safety, and process economics, making reliable dynamic predictions indispensable for industrial design, optimization, and control~\cite{FANG2025101152, jackson2025ten}. Accordingly, this work focuses on radical polymerization as a representative and industrially relevant polymerization technique.

Mechanistic models based on first-principles chemical kinetics have long served as the foundation of polymer reaction engineering~\cite{noble2013first}. These models provide physical interpretability and mechanistic insight into polymerization pathways. Radical polymerization is typically described as following the steps of initiation, propagation, and termination, sequentially adding monomers to growing chains. Both deterministic and stochastic formulations have been used~\cite{FANG2025101152}. Deterministic modeling is commonly used in reactor-scale analysis and process optimization to describe macroscopic behaviors such as conversion and molecular weight~\cite{jones2023reversed}. However, these models typically rely on simplifying assumptions and lack detailed molecular resolution, particularly with respect to chain length–dependent reactivity and sequence effects~\cite{jones2023reversed}.
Stochastic models are typically used to describe molecular-level polymerization behavior, such as chain-length distributions and chain topology, whereas deterministic models are more commonly applied to reactor-scale process analysis and optimization~\cite{wulkow2021deterministic}. The limitations of both approaches have motivated growing interest in machine learning (ML)-based modeling approaches~\cite{schweidtmann2020machine,ge2025machine, VONSTOSCH201486}.

The body of ML literature in polymer science has expanded rapidly in recent years. In the following, we focus on ML approaches relevant to dynamic polymerization process modeling under limited data availability.

Most ML contributions in the literature focus on predicting static kinetic quantities, such as propagation rate coefficients or comonomer reactivity ratios, directly from molecular structure, often using SMILES-based descriptors~\cite{wang2024machine}, molecular fingerprints~\cite{shi2022supervised}, or graph-based representations~\cite{li2024gatboost,inae2025modeling}. Recent work has also explored ML-based approaches for polymerization process modeling, including the prediction of time-evolving quantities such as molecular weight distributions and other reactor-scale properties~\cite{https://doi.org/10.1002/cjce.25635,BARDOOLI20211}.
Related data-driven efforts extend beyond kinetics, for example, to polymer-solvent compatibility, where hybrid symbolic-continuous ML models have been used to jointly learn solvation states and quantitative solubility behavior from experimental data~\cite{liew2025learning}. Ge et al.~\cite{ge2025machine} review the growing role of ML in polymer research, including applications relevant to polymerization, and highlight its potential to accelerate property prediction and materials discovery. However, the authors also emphasize several key limitations of purely data-driven approaches, such as the scarcity and inconsistency of polymerization data, the dominance of small-data regimes, difficulties in representing polymerization mechanisms and reaction dynamics, limited extrapolation beyond training conditions, and the lack of physical interpretability. As a result, Ge et al.~\cite{ge2025machine} argue that ML models that neglect polymer theory and reaction kinetics risk learning correlations rather than causative behavior. The authors therefore advocate theory-guided ML, where first-principles knowledge is integrated with data-driven models to improve robustness, interpretability, and generalization. These observations, together with the challenges of modeling polymerization dynamics under sparse and noisy data conditions, further motivate the hybrid modeling approach pursued in this work.

The integration of mechanistic and data-driven models has been studied for decades under the framework of hybrid modeling~\cite{hybrid2002,hybrid2008,hybrid1992}. In recent years, renewed interest in these approaches has emerged in the context of modern ML methods and differentiable dynamical systems~\cite{merkelbach2022hybridml,karniadakis2021physics}. Multiple strategies for physics-informed and theory-guided ML approaches have been proposed~\cite{audus2017polymer,lin2022deep,ethier2023integrating,li2025theory, VELIOGLU2025108899}. One approach incorporates theoretical knowledge directly into the modeling framework, as shown by Audus et al.~\cite{audus2022leveraging}, where combining polymer scaling theory with simulation data improved predictions even when the theoretical model was approximate. Related studies relevant to polymerization and processing demonstrate that physically meaningful parameters, such as interaction parameters or kinetic descriptors, can act as informative metafeatures, enabling accurate predictions under limited data availability~\cite{ethier2023integrating}. Beyond such representations, theory can also be embedded into model architectures and training procedures; for example, physics-guided neural networks with physically motivated activation functions and constraints have been shown to significantly reduce prediction errors in polymer processing and thermal response problems~\cite{zobeiry2021theory}. Collectively, these studies highlight that theory-guided ML provides a promising framework for polymerization modeling.

For the batch free-radical polymerization (FRP) systems considered in this work, the process dynamics are described by systems of ordinary differential equations derived from reaction kinetics. Building on this structure, we formulate a hybrid Neural Ordinary Differential Equation (NODE) model for FRP dynamics under limited data availability. NODEs extend residual neural networks to the continuous-time setting by parameterizing the system dynamics through ordinary differential equations~\cite{chen2018neural,NEURIPS2020_293835c2}. Recent work in chemical engineering has demonstrated that NODE-based models provide an effective framework for reaction systems under partial mechanistic knowledge and sparse data~\cite{sorourifar2023physics}. NODEs also inherit favorable properties such as parameter efficiency and robustness to perturbations~\cite{yan2022robustnessneuralordinarydifferential}. Their continuous-time structure provides a natural framework for combining mechanistic balance equations with learned dynamic contributions within the same dynamical system. While NODE-based approaches have attracted growing interest in chemical engineering and scientific machine learning, their application to hybrid polymerization modeling remains comparatively limited~\cite{ge2025machine}, motivating the approach adopted in this work, where we build on this framework by retaining the known polymerization kinetics explicitly and introducing neural networks only for the missing or approximated dynamic contributions.

In this work, FRP serves both as the application focus and as a representative example of a reaction system with partially-characterized kinetics. The proposed hybrid NODE formulation follows a structured modeling strategy in which mechanistic relations that are well established, broadly valid, and readily parameterized are retained explicitly, while difficult-to-model or uncertain kinetic contributions are learned from data. In the present polymerization system, the mechanistic mass balances and known kinetic dependencies are preserved, whereas the effective radical contribution governing monomer consumption is represented through a learned closure term. The neural-network component is therefore not used to replace the mechanistic model but rather to complement it where the kinetics are difficult to specify accurately within the reduced mechanistic formulation. This formulation preserves the physical structure of the system while reducing the learning problem compared to a fully data-driven approach and reducing the amount of data required for training, since only the unresolved dynamic contribution is learned from data. 

By combining mechanistic modeling with data-driven learning in a structure-consistent continuous-time formulation, this work proposes a hybrid NODE approach for FRP systems. The resulting model achieves improved predictive accuracy and generalization across operating conditions, particularly in sparse-data regimes where only limited measurements are available.

The remainder of this work is organized as follows. First, the mechanistic model used in this work for the polymerization process and its state-space formulation are presented in Section~\ref{sec:frp_kinetics}. 
Next, Section~\ref{sec:ml_dynamic_models} discusses the ML formulations considered for dynamic process modeling, followed by the proposed hybrid Neural ODE formulation in Section~\ref{sec:hybrid_NODE}.
Section~\ref{sec:implementation} then presents the model implementation, training procedure, and evaluation scenarios.
Finally, Section~\ref{sec:results} presents and discusses the results, and Section~\ref{Conclusion and Outlook} concludes the paper with a summary and outlook.

\section{Free-Radical Polymerization Process Modeling}
\label{sec:frp_kinetics}
We consider a homogeneous batch FRP reactor operated under constant-volume conditions. This section summarizes the standard mechanistic model used as the underlying mechanistic structure throughout this work. The model serves two purposes. First, it is used to generate synthetic data for the present case study, which serves as a proof of concept for the proposed approach; in a real-world implementation the data would come from experiments. Second, it provides a partial mechanistic description of the polymerization dynamics that is incorporated into the hybrid NODE formulation, and it is presented here for completeness and to make the subsequent hybrid NODE formulation self-contained. First, the key aspects of FRP kinetics and the corresponding material balances are summarized. The resulting species balances are subsequently combined with the kinetic rate expressions and approximated using a finite-dimensional chain-length discretization, yielding a continuous-time state-space representation of the polymerization dynamics.

\subsection{Free-Radical Polymerization Kinetics and Modeling}
\label{sec:pivot_discretization}
Modeling polymerization kinetics requires combining well-established mechanistic reaction pathways with kinetic contributions that are often difficult to specify or measure~\cite{10.1071/CH02114}. This subsection focuses on the kinetic aspects of the FRP process, with limited emphasis on detailed chemical considerations. The model formulation follows the reaction network modeling workflow described in~\cite{WALZ2017143}, i.e., first defining the reaction network, then deriving the species balances using stoichiometric relations, and finally imposing constitutive kinetic rate expressions. Despite the availability of general-purpose mechanistic models for FRP, their application to specific polymerization systems remains challenging due to the complexity of the kinetic rate, including propagation, termination, and chain-transfer reactions, and the need for substantial kinetic characterization and parameter estimation. The mechanistic model introduced here, therefore, serves as the underlying mechanistic structure for the subsequent hybrid formulation.

A key feature of FRP is the presence of an adjustable radical source, typically provided by a thermolabile initiator that decomposes upon heating to generate free radicals, i.e., highly reactive species with an unpaired electron. During the polymerization of monomers, each monomer addition preserves the radical structure at the chain end, allowing the growing chain to remain active and continue reacting with additional monomer units. The reaction system contains the following species:
\[
\mathcal{S} = \{ I,\, M,\, CTA,\, S \} \cup \{ R_n^\bullet,\, D_n \}_{n\ge1}
\]
where, $I$ denotes the initiator, $M$ the monomer, $CTA$ the chain transfer agent, and $S$ the solvent. The species $R_n^\bullet$ represents the living radical of chain length $n$, i.e., a polymer chain carrying an active radical center capable of further propagation, while $D_n$ denotes the dead polymer chain of length $n$, i.e., a chain that no longer contains an active radical center as a result of termination reactions.
The conventional kinetic description of FRP comprises initiation, propagation, and termination, with additional chain-transfer reactions included to account for transfer to chain transfer agent, monomer, and solvent~\cite {matyjaszewski2002handbook,doi:https://doi.org/10.1002/9781119820123.ch3,10.1039/9781849737425-00001}. In the following, $n,m \in \mathbb{N}_{\ge 1}$ denote polymer chain lengths, $f \in (0,1]$ denotes the initiator efficiency, i.e., the fraction of decomposed initiator radicals that successfully initiate polymer chains, and $k_d$, $k_p$, $k_{tr}$, $k_{trm}$, $k_{trs}$, $k_{td}$, and $k_{tc}$ the kinetic rate constants associated with initiator decomposition, propagation, chain transfer, and termination, respectively. The corresponding reaction scheme is given by:
\begin{align*}
\text{Initiator decomposition:}
&\quad I &&\xrightarrow{f k_d}&& 2\,R_1^\bullet \\[6pt]
\text{Propagation:}
&\quad R_n^\bullet + M &&\xrightarrow{k_p}&& R_{n+1}^\bullet \\[6pt]
\text{Chain transfer to chain transfer agent:}
&\quad R_n^\bullet + CTA &&\xrightarrow{k_{tr}}&& D_n + R_1^\bullet \\
\text{Chain transfer to monomer:}
&\quad R_n^\bullet + M &&\xrightarrow{k_{trm}}&& D_n + R_1^\bullet \\
\text{Chain transfer to solvent:}
&\quad R_n^\bullet + S &&\xrightarrow{k_{trs}}&& D_n + R_1^\bullet \\[6pt]
\text{Termination by disproportionation:}
&\quad R_n^\bullet + R_m^\bullet &&\xrightarrow{k_{td}}&& D_n + D_m \\
\text{Termination by combination:}
&\quad R_n^\bullet + R_m^\bullet &&\xrightarrow{k_{tc}}&& D_{n+m}
\end{align*}
For simplicity, the intermediate radical species generated during chain-transfer reactions to monomer ($M$), chain transfer agent ($CTA$), and solvent ($S$) are not modeled explicitly; instead, each transfer event is represented by the formation of an effective propagating radical species $R_1^\bullet$~\cite{moad2005chemistry}.
For the considered homogeneous batch reactor with constant volume $V$, the material balances are written as:
\[\left.\frac{dc_i}{dt}\right|_{t}= R_i(t)\]
where $c_i$ denotes the concentration of species $i$ and $R_i$ the net rate of formation. The species production rates are expressed as

\[R_i(t) = \sum_{j \in \mathcal{R}} \nu_{i,j} r_j(t),\]
where $r_j(t)$ denotes the reaction rate associated with reaction $j \in \mathcal{R}$ and $\nu_{i,j}$ the corresponding stoichiometric coefficient, where $\mathcal{R}$ denotes the set of reactions listed in the reaction stoichiometry scheme above. The reaction rates are modeled as
\[
r_j(t) = k_j(T)\, f_j(\mathbf{c}(t)),
\]
where $f_j(\mathbf{c}(t))$ denotes the concentration-dependent kinetic term, which, under the assumption of mass-action kinetics, is given by the product of reagent concentrations raised to powers corresponding to the reaction stoichiometry. The  temperature dependence is modeled using Arrhenius form:
\[
k_j(T) = k_{0,j}\exp\left(-\frac{E_j}{RT}\right).
\]
For compactness, the temperature dependence of the rate constants is suppressed in the following, i.e., \(k_j \equiv k_j(T)\).
The elementary reaction rates corresponding to the kinetic scheme are defined as
\begin{align*}
r_d(t)
&= f k_d\, c_I(t), \\
r_{p,n}(t)
&= k_p\, c_M(t)\, c_{R_n^\bullet}(t),
&\qquad n \ge 1, \\
r_{tr,n}(t)
&= k_{tr}\, c_{CTA}(t)\, c_{R_n^\bullet}(t),
&\qquad n \ge 1, \\
r_{trm,n}(t)
&= k_{trm}\, c_M(t)\, c_{R_n^\bullet}(t),
&\qquad n \ge 1, \\
r_{trs,n}(t)
&= k_{trs}\, c_S(t)\, c_{R_n^\bullet}(t),
&\qquad n \ge 1, \\
r_{td,n,m}(t)
&= k_{td}\, c_{R_n^\bullet}(t)\, c_{R_m^\bullet}(t),
&\qquad n,m \ge 1, \\
r_{tc,n,m}(t)
&= k_{tc}\, c_{R_n^\bullet}(t)\, c_{R_m^\bullet}(t),
&\qquad n,m \ge 1.
\end{align*}
For the molecular species, the balances read
\begin{align*}
\left.\frac{d c_I}{dt}\right|_{t}
&= -\,\frac{r_d(t)}{f}, \\
\left.\frac{d c_M}{dt}\right|_{t}
&= -\sum_{n\ge1} r_{p,n}(t) - \sum_{n\ge1} r_{trm,n}(t), \\
\left.\frac{d c_{CTA}}{dt}\right|_{t}
&= -\sum_{n\ge1} r_{tr,n}(t), \\
\left.\frac{d c_S}{dt}\right|_{t}
&= -\sum_{n\ge1} r_{trs,n}(t).
\end{align*}
where $f$ is the initiator efficiency and $r_d(t)= f k_d\, c_I(t)$ denotes the effective rate of polymerizing radical generation. While only a fraction $f$ of the generated primary radicals become active in polymerization, the initiator decomposes at rate $k_dc_I(t)$. Consequently, the initiator consumption is governed by the decomposition rate.
For the living radical of chain length $n=1$, the balance is
\begin{align*}
\left.\frac{d c_{R_1^\bullet}}{dt}\right|_{t}
&= 2r_d(t)
- r_{p,1}(t)
- r_{tr,1}(t)
- r_{trm,1}(t)
- r_{trs,1}(t) \nonumber\\
&\quad
+ \sum_{n\ge1} r_{tr,n}(t)
+ \sum_{n\ge1} r_{trm,n}(t)
+ \sum_{n\ge1} r_{trs,n}(t) \nonumber\\
&\quad
- \sum_{m\ge1} r_{td,1,m}(t)
- \sum_{m\ge1} r_{tc,1,m}(t).
\end{align*}
For living radicals of chain length $n\ge2$, the balances are
\begin{align*}
\left.\frac{d c_{R_n^\bullet}}{dt}\right|_{t}
&= r_{p,n-1}(t)
- r_{p,n}(t)
- r_{tr,n}(t)
- r_{trm,n}(t)
- r_{trs,n}(t) \nonumber\\
&\quad
- \sum_{m\ge1} r_{td,n,m}(t)
- \sum_{m\ge1} r_{tc,n,m}(t),
\qquad n\ge2.
\end{align*}
For dead polymer chains, the balances are
\begin{align*}
\left.\frac{d c_{D_n}}{dt}\right|_{t}
&= r_{tr,n}(t)
+ r_{trm,n}(t)
+ r_{trs,n}(t) \nonumber\\
&\quad
+ \sum_{m\ge1} r_{td,n,m}(t)
+ \frac{1}{2}\sum_{m=1}^{n-1} r_{tc,m,n-m}(t),
\qquad n\ge1.
\end{align*}
The balances above define the formal infinite-dimensional population balance model. In principle, this leads to a system of population balances over the chain length, where the concentrations of living and dead polymer species are defined for all chain lengths $n$. However, directly resolving these balances is computationally demanding, as propagation continuously shifts the distribution and termination by combination generates products at chain length $n+m$.
To obtain a numerically tractable formulation, we adopt a pivot-based discretization of the chain-length domain, following the approach of Butté et al.~\cite{butte2002evaluation1,butte2002evaluation2}. In the implemented reference model, the infinite-dimensional balances are approximated on a finite pivot grid, and the pairwise termination interactions are closed through the total radical concentration. In this method, the chain-length distribution is represented on a finite set of pivot points, and the evolution of the distribution is approximated by projecting reaction products onto these pivots. This allows the computation of chain-length distributions while maintaining a manageable number of state variables.

In the implemented mechanistic reference model, the infinite-dimensional chain-length distributions are discretized on a finite pivot grid following the method of Butté et al.~\cite{butte2002evaluation1,butte2002evaluation2}. After discretization, the same kinetic rate expressions are evaluated for the finite set of pivot species \(i=1,\dots,N_p\), with the total radical concentration approximated by $c_{R^\bullet}(t)=\sum_i c_{R_i^\bullet}(t).$
The termination rates are then expressed in terms of this total radical concentration, resulting in a closure of the pairwise radical interactions~\cite{closure}. The corresponding species balances read
\begin{align*}
\left.\frac{d c_I}{dt}\right|_{t}
&= -\, k_d\, c_I(t), \\[6pt]
\left.\frac{d c_M}{dt}\right|_{t}
&= -\,k_p\, c_M(t)\, c_{R^\bullet}(t)
   - k_{trm}\, c_M(t)\, c_{R^\bullet}(t), \\[6pt]
\left.\frac{d c_{CTA}}{dt}\right|_{t}
&= -\,k_{tr}\, c_{CTA}(t)\, c_{R^\bullet}(t), \\[6pt]
\left.\frac{d c_S}{dt}\right|_{t}
&= -\,k_{trs}\, c_S(t)\, c_{R^\bullet}(t).
\end{align*}
For the living radicals on the pivot grid, the balances are given by
\begin{align*}
\left.\frac{d c_{R_i^\bullet}}{dt}\right|_{t}
&= \delta_{i1}\,2 f k_d\, c_I(t)
+ \mathcal{P}_i\!\left(c_{R^\bullet}(t),c_M(t)\right) \nonumber\\
&\quad
- \left(
k_{tr}\,c_{CTA}(t)
+ k_{trm}\,c_M(t)
+ k_{trs}\,c_S(t)
\right)c_{R_i^\bullet}(t) \nonumber\\
&\quad
- \left(k_{td} + k_{tc}\right)
c_{R_i^\bullet}(t)\, c_{R^\bullet}(t),
\qquad i=1,\dots,N_p
\end{align*}
For the dead polymer distribution, the balances read
\begin{align*}
\left.\frac{d c_{D_i}}{dt}\right|_{t}
&=
\left(
k_{tr}\,c_{CTA}(t)
+ k_{trm}\,c_M(t)
+ k_{trs}\,c_S(t)
\right)c_{R_i^\bullet}(t) \nonumber\\
&\quad
+ k_{td}\,c_{R_i^\bullet}(t)\,c_{R^\bullet}(t)
+ \mathcal{C}_i\!\left(c_{R^\bullet}(t)\right),
\qquad i=1,\dots,N_p
\end{align*}
The operator $\mathcal{P}_i(\cdot)$ represents the propagation-induced shift of radicals between neighboring pivots, and $\mathcal{C}_i(\cdot)$ denotes the projection of combination products onto the pivot grid. Both operators follow the pivot-based discretization procedure introduced by Butté et al.~\cite{butte2002evaluation1,butte2002evaluation2}.
The resulting mechanistic model provides a structured description of the polymerization dynamics. In the following, this model is reformulated in a continuous-time state-space representation to facilitate its integration with the hybrid NODE formulation introduced later.
\subsection{State-Space Modeling in Continuous Time}
The full radical and polymer chain-length distributions are discretized using the pivot method introduced in Section~\ref{sec:pivot_discretization}. Following the implementation used for data generation and repeated model evaluations during training, the chain-length distributions are approximated using 16 pivots, providing a practical compromise between numerical resolution and computational effort.
The resulting system states, therefore, include the molecular species concentrations together with the discretized living radical and dead-polymer populations.
To formulate the mechanistic reference model in state-space form, we collect all dynamic variables into the state vector
\[
\mathbf{x}_{\mathrm{mech}}(t)=
\begin{bmatrix}
c_I(t) &
c_M(t) &
c_{CTA}(t) &
c_S(t) &
\mathbf{c}_{R^\bullet}(t)^{\top} &
\mathbf{c}_{D}(t)^{\top}
\end{bmatrix}^{\top}
\]
with
\[
\mathbf{c}_{R^\bullet}(t)=
\begin{bmatrix}
c_{R_1^\bullet}(t) & \cdots & c_{R_{16}^\bullet}(t)
\end{bmatrix}^{\top},
\qquad
\mathbf{c}_{D}(t)=
\begin{bmatrix}
c_{D_1}(t) & \cdots & c_{D_{16}}(t)
\end{bmatrix}^{\top}.
\]
The reactor is assumed to operate under constant-volume conditions. Consequently, $V$ is a constant parameter rather than a time-dependent state, and all states are expressed in terms of concentrations.
The initial conditions of the system are given by
\[
\mathbf{x}_{\mathrm{mech}}(0)=
\begin{bmatrix}
c_{I,0} &
c_{M,0} &
c_{CTA,0} &
c_{S,0} &
0 &
\cdots &
0
\end{bmatrix}^{\top}
\]
since initially, no radicals or polymer chains are present.

The kinetic parameters of the mechanistic model are collected in the parameter vector
\[
\mathbf{p}_{\mathrm{mech}}=
\begin{bmatrix}
k_d &
k_p &
k_{tr} &
k_{trm} &
k_{trs} &
k_{td} &
k_{tc} &
f
\end{bmatrix}^{\top}
\]
where the parameters $k_j$ denote effective, temperature-dependent rate constants, while $f \in (0,1]$ denotes the initiator efficiency.
The operating conditions are represented by the input vector
\[
\mathbf{u}_{\mathrm{mech}}(t)=
\begin{bmatrix}
T
\end{bmatrix},
\]
where $T$ denotes the reactor temperature, which affects the reaction rates through Arrhenius-type dependencies.

The system dynamics can thus be written in continuous-time state-space form as
\[
\dot{\mathbf{x}}_{\mathrm{mech}}(t) =
\mathbf{f}_{\mathrm{mech}}
\left(
\mathbf{x}_{\mathrm{mech}}(t),
\mathbf{u}_{\mathrm{mech}}(t),
\mathbf{p}_{\mathrm{mech}}
\right),
\]
where $\mathbf{f}_{\mathrm{mech}}(\cdot)$ represents the mechanistic reaction kinetics defined by the species balances.

The model output corresponds to the monomer conversion,
\[
y_{\mathrm{mech}}(t)
=
1 - \frac{c_M(t)}{c_{M,0}}.
\]
This formulation provides the structural foundation for the hybrid modeling approach introduced in the following section.

\section{Machine Learning for Dynamic Process Modeling}
\label{sec:ml_dynamic_models}

This section introduces the ML formulations considered in this work for modeling dynamic process behavior from time-series data. The formulations are presented to establish the continuous- and discrete-time learning frameworks used later for comparison with the proposed hybrid NODE approach and to provide the notation and modeling structure required for the subsequent hybrid formulations. We first define the general learning problem for nonlinear dynamical systems and the associated training objective. We then present two model classes considered throughout this work: discrete-time neural transition models and continuous-time Neural Ordinary Differential Equations (NODEs).

\subsection{Learning Problem and Training Objective}
\label{sec:learning_problem}

We consider a general non-autonomous nonlinear dynamic system in continuous time~\cite{kloeden2011nonautonomous}:
\begin{equation}
    \dot{\mathbf{x}}(t)=\mathbf{f}\big(\mathbf{x}(t),\mathbf{u}(t)\big),
    \qquad
    \mathbf{x}(t_0)=\mathbf{x}_0,
\end{equation}
where $\mathbf{x}(t)\in\mathbb{R}^{n_x}$ denotes the state vector, $\mathbf{u}(t)\in\mathbb{R}^{n_u}$ the input vector, and $\mathbf{f}$ the nonlinear state-transition function.

In general, the full state vector is not directly observable. Instead, measurements are obtained through
\begin{equation}
    \mathbf{y}(t)=\mathbf{h}\big(\mathbf{x}(t)\big)+\boldsymbol{\varepsilon}(t),
    \qquad
    \boldsymbol{\varepsilon}(t)\sim\mathcal{N}(0,\Sigma),
\end{equation}
where $\mathbf{y}(t)\in\mathbb{R}^{n_y}$ denotes the system outputs and $\boldsymbol{\varepsilon}(t)\in\mathbb{R}^{n_y}$ represents measurement noise. 

At discrete sampling times $t_i$, the measured outputs are denoted by
\[
\mathbf{y}_i := \mathbf{y}(t_i) + \boldsymbol{\varepsilon}_i.
\]
Given time-series data
\[
\mathcal{D}=\left\{\mathbf{u}(t_i),\mathbf{y}_i\right\}_{i=1}^{N},
\]
the learning problem consists of estimating the trainable parameter vector of the neural network $\boldsymbol{\theta}$ such that the model predictions match the observed measurements. This is formulated as the following training objective:
\begin{equation}
\label{eq:training}
\min_{\theta}\;
\frac{1}{N}\sum_{i=1}^{N}
\left\|
\hat{\mathbf{y}}_{\theta}(t_i)-\mathbf{y}_i
\right\|_2^2
+\lambda_{\mathrm{reg}}\mathcal{R}(\theta),
\end{equation}
where $\hat{\mathbf{y}}_{\boldsymbol{\theta}}(t_i)$ denotes the model prediction at time $t_i$, and $\mathcal{R}(\boldsymbol{\theta})$ is a regularization term, e.g., an $\ell_2$ penalty on the network weights, weighted by $\lambda_{\mathrm{reg}} \ge 0$.

Depending on the model class, we obtain predictions either by recursive discrete-time updates or by solving a continuous-time initial-value problem.

\subsection{Discrete-Time Neural Modeling}
\label{sec:discrete_model}

A general discrete-time approximation of a nonlinear dynamical system can be written as 
\begin{equation}
    \mathbf{x}_{k+1}=\mathbf{F}(\mathbf{x}_k,\mathbf{u}_k),
\end{equation}
where $\mathbf{x}_k \in \mathbb{R}^{n_x}$ denotes the system state at time step $k$, 
$\mathbf{u}_k \in \mathbb{R}^{n_u}$ the corresponding input, and 
$\mathbf{F}: \mathbb{R}^{n_x}\times\mathbb{R}^{n_u}\rightarrow\mathbb{R}^{n_x}$ 
the discrete-time state-transition mapping~\cite{WANG19921}.
When $\mathbf{F}$ is represented by a neural network with parameters $\theta$, this yields:
\begin{equation}
    \mathbf{x}_{k+1}=\mathbf{F}_{\boldsymbol{\theta}}^{\mathrm{disc}}\big(\mathbf{x}_k,\mathbf{u}_k,\Delta t_k\big),
    \label{eq:discrete_nn}
\end{equation}
where $\boldsymbol{\theta}$ denotes the trainable parameters of the neural network. Here, $\Delta t_k$ is included to account for variable time steps, enabling learning from irregularly sampled data. This formulation is purely data-driven and treats the dynamics as a static regression problem that maps $(\mathbf{x}_k,\mathbf{u}_k,\Delta t_k)$ to $\mathbf{x}_{k+1}$ at each time step.  
While simple to implement, it does not explicitly enforce an underlying continuous-time structure, and its performance depends strongly on the chosen time discretization and the density of available training data, typically requiring sufficiently dense time-series measurements to accurately capture fast dynamics and avoid instability in recursive predictions~\cite{WANG19921}.

\subsection{Neural Ordinary Differential Equations}

Neural Ordinary Differential Equations (NODEs) parameterize the continuous-time dynamics directly~\cite{massaroli2020dissecting}:
\begin{equation}
    \dot{\mathbf{x}}(t) = \mathbf{f}_{\boldsymbol{\theta}}^{\mathrm{NODE}}\big(\mathbf{x}(t),\mathbf{u}(t)\big),
    \label{eq:node_basic}
\end{equation}
where $\mathbf{x}(t) \in \mathbb{R}^{n_x}$ denotes the system state, 
$\mathbf{u}(t) \in \mathbb{R}^{n_u}$ the input vector, and 
$\mathbf{f}_{\boldsymbol{\theta}}^{\mathrm{NODE}}: \mathbb{R}^{n_x} \times \mathbb{R}^{n_u} \rightarrow \mathbb{R}^{n_x}$ 
is a neural network with trainable parameters $\theta$ that approximates the continuous-time state-transition function.

We obtain the state trajectory by solving the corresponding initial-value problem with initial condition $\mathbf{x}(t_0)=\mathbf{x}_0$:
\begin{equation}
    \mathbf{x}(t)
    = \mathbf{x}_0
    + \int_{t_0}^{t}
        \mathbf{f}_{\boldsymbol{\theta}}^{\mathrm{NODE}}\big(\mathbf{x}(\tau),\mathbf{u}(\tau)\big)\,d\tau,
    \label{eq:node_integral}
\end{equation}
We train the model parameters by minimizing the objective in Equation~\eqref{eq:training}, where we obtain predicted trajectories by solving the corresponding initial-value problem. Gradients are computed via backpropagation through the ODE solver, typically using adjoint sensitivity methods. For further details on NODE training and adjoint-based methods, we refer to~\cite{chen2018neural}.

This formulation treats the neural network as a parameterization of the continuous-time state-transition function and provides a flexible framework for learning dynamics directly from time-series data. It forms the foundation for the hybrid modeling approach introduced in the following section. In contrast to discrete-time models, NODEs directly model the time derivative of the state and reconstruct trajectories through integration using a differentiable ODE solver. As a result, the model enforces a continuous-time dynamic structure and can naturally accommodate irregularly sampled measurements~\cite{chen2018neural}.

\section{Hybrid Neural Ordinary Differential Equations}
\label{sec:hybrid_NODE}
As discussed in the Introduction, purely data-driven models for polymerization dynamics are difficult to train reliably under limited and noisy data conditions~\cite{jackson2025ten}. At the same time, the overall structure of FRP dynamics is characterized through well-established reaction mechanisms and mass balances, as presented in Section~\ref{sec:frp_kinetics}, while several constitutive kinetic contributions remain uncertain or difficult to model accurately. This mechanistic structure can therefore be incorporated directly into the model formulation to reduce the amount of information that must be inferred from data alone. In the following, we first introduce the general hybrid NODE framework and then subsequently specialize it to the FRP system considered in this work.

\subsection{Hybrid NODE Formulation}
\label{sec:hybrid_general}
Our approach is to retain the known mechanistic structure of the polymerization system and introduce data-driven learning only for the unresolved kinetic contribution required to complete the reduced mechanistic model. In particular, the mechanistic mass balances and established kinetic relations are retained explicitly, while the effective radical concentration governing monomer consumption is approximated through a neural-network surrogate. The neural component, therefore, does not replace the full mechanistic model but instead complements it by learning only the part of the dynamics that is difficult to characterize accurately within the reduced mechanistic formulation. Unlike conventional hybrid correction approaches, the proposed formulation does not learn additive corrections to an existing mechanistic model but instead learns the unresolved closure relation required to complete an otherwise structurally consistent mechanistic description. Here, the term “closure” refers to the representation of an unresolved quantity required to complete the mechanistic description of the reduced system dynamics.

To implement this approach, we employ a hybrid Neural Ordinary Differential Equation (NODE) formulation. This choice is motivated by the fact that polymerization processes are naturally described by continuous-time ordinary differential equations derived from reaction kinetics, as shown in Section~\ref{sec:frp_kinetics}. NODEs provide a compatible framework in which the system dynamics are learned directly in continuous time, allowing mechanistic and learned components to be embedded within the same system of differential equations.

Within this framework, the system dynamics are written as
\begin{equation}
\dot{\mathbf{x}}(t)
=
\mathbf{f}_{\mathrm{phys}}\!\left(
\mathbf{x}(t),
\mathbf{u}(t),
\mathbf{f}_{\boldsymbol{\theta}}^{\mathrm{hyb}}\big(\mathbf{x}(t),\mathbf{u}(t)\big)
\right),
\label{eq:hybrid_general}
\end{equation}
where $\mathbf{f}_{\mathrm{phys}}$ represents the mechanistic balance equations and $\mathbf{f}_{\boldsymbol{\theta}}^{\mathrm{hyb}}$ approximates the unresolved constitutive contribution required to close the system. The continuous-time formulation preserves consistency with the underlying reaction kinetics while embedding the learned closure relation directly within the governing differential equations.

In the following, this formulation is specialized to the FRP system considered in this work.
\subsection{Hybrid NODE Formulation for Free-Radical Polymerization Dynamics}
\label{sec:hybrid_frp}

We now specify the general hybrid NODE formulation introduced in Section~\ref{sec:hybrid_general} to the FRP system described in Section~\ref{sec:frp_kinetics}.
Based on the reaction mechanism, monomer is consumed primarily through propagation and, to a lesser extent, through chain transfer to monomer~\cite{10.1071/CH02114}. The monomer balance can therefore be written as:
\begin{equation}
\left.\frac{d c_M}{dt}\right|_{t}
=
-\left(k_p+k_{trm}\right)c_M(t)c_{R^\bullet}(t),
\label{eq:monomer_balance_mech}
\end{equation}
\begin{equation}
\left.\frac{d c_M}{dt}\right|_{t}
\approx
-\, k_p\, c_M(t)\, c_{R^\bullet}(t),
\label{eq:monomer_approx}
\end{equation}
where $c_{R^\bullet}(t)=\sum_i c_{R_i^\bullet}(t)$ denotes the total radical concentration.

Following the hybrid modeling strategy, we retain this balance structure and focus on closing the term that is difficult to describe accurately, namely the total radical concentration. This quantity results from the coupled effects of initiation, propagation, transfer, and termination reactions, and its evaluation in the full mechanistic model depends on the discretized radical population balance. We reformulate the monomer balance using the logarithmic state~\cite{10.1071/CH02114}
\begin{equation}
z(t)=\log\!\left(\frac{c_M(t)}{c_{M,0}}\right).
\end{equation}
In particular, the total radical concentration is represented by a neural surrogate,
\begin{equation}
c_{R^\bullet}(t)
\;\approx\;
\widehat{R}_{\theta}\big(z(t), \mathbf{u}\big),
\end{equation}
where $\widehat{R}_{\theta}$ denotes a learned approximation that closes the monomer balance, $z(t)$ denotes the logarithmic state, and $\mathbf{u}$ represents the operating conditions.

The resulting hybrid monomer balance becomes
\begin{equation}
\left.\frac{d c_M}{dt}\right|_{t}
=
-\, k_p\, c_M(t)\,
\widehat{R}_{\theta}\big(z(t), \mathbf{u}\big),
\label{eq:monomer_balance_hybrid}
\end{equation}

Applying the chain rule, we obtain
\begin{equation}
\dot z(t)
=
\frac{1}{c_M(t)}
\left.\frac{d c_M}{dt}\right|_{t}.
\end{equation}

Substituting the hybrid monomer balance yields
\begin{equation}
\dot z(t)
=
-\,k_p\,
\widehat{R}_{\theta}\big(z(t),\mathbf{u}\big),
\qquad
y(t)=1-\exp\big(z(t)\big),
\label{FRP-hyb}
\end{equation}

The neural surrogate $\widehat{R}_{\theta}$ is embedded directly within the NODE formulation and trained end-to-end through the governing differential equations.

In summary, the proposed formulation combines mechanistic balance equations with a learned closure relation for the radical concentration, yielding a structure-consistent hybrid NODE model with the flexibility required to capture complex reactions such as FRP dynamics under limited data availability.

\section{Model Implementation and Training}
\label{sec:implementation}
This section describes the implementation of the mechanistic, discrete-time, NODE, and hybrid NODE models introduced in Sections~\ref{sec:frp_kinetics},~\ref{sec:ml_dynamic_models}, and~\ref{sec:hybrid_NODE}, together with the data generation procedure, evaluation scenarios, and training methodology used to assess their performance for FRP dynamics.

\subsection{Reference Trajectories and Data Generation}

We generate all training and evaluation data using the mechanistic simulator introduced in Section~\ref{sec:frp_kinetics} for the FRP of methyl methacrylate (MMA), using the pivot discretization approach of Butté and Morbidelli~\cite{butte2002evaluation1,butte2002evaluation2}.  
The simulator provides time-resolved concentration trajectories for the molecular species, from which the monomer conversion is computed as~\cite{tan2022machine}
\begin{align}
X(t) = 1 - \frac{c_M(t)}{c_{M,0}}.
\end{align}
Each trajectory consists of 25 normalized time points over a fixed batch duration of three hours. To evaluate model performance under limited-data conditions, we use only the first 10 time points for training, while the remaining 15 points constitute the evaluation segment. We apply this split consistently across all scenarios.

\subsection{Evaluation Scenarios}
The evaluation scenarios examine the three model formulations under progressively more challenging data conditions. In all scenarios, we keep the model structures and loss formulations consistent across models. Only the data sampling pattern and recipe variability change.

\paragraph{Scenario 1: Limited data with regular sampling.}
In the first scenario, we consider a single trajectory on a regular time grid with 25 normalized time points. We train all models using only the first 10 time points and evaluate predictions over the full 25-point horizon. This setting isolates the ability of each model to reconstruct a smooth conversion trajectory from sparse but regularly sampled data.

\paragraph{Scenario 2: Limited data with irregular sampling.}
In the second scenario, we consider the same limited-data setting, but we replace the regular grid with irregular time points. We keep the number of training points fixed at 10 and evaluate over the full trajectory. By changing only the sampling pattern, we assess how sensitive each model is to nonuniform measurement intervals while keeping the data quantity and training procedure unchanged. This experiment evaluates robustness to irregular measurement schedules, which are common in practical polymerization experiments \cite{RICHARDS20061447}.

\paragraph{Scenario 3: Generalization under noisy data.}
In the third scenario, we evaluate generalization to unseen operating conditions in the presence of measurement noise. We generate full conversion trajectories using the mechanistic simulator and add Gaussian noise with standard deviation $\sigma = 0.02$ for the conversion observations.
We generate training trajectories by randomly sampling temperature, initial monomer concentration, initiator concentration, chain-transfer agent concentration, and solvent concentration over representative ranges.
In this scenario, the models are trained on a limited set of complete noisy trajectories and then predict the conversion dynamics for completely unseen recipes. This setting tests the ability of each formulation to preserve physically consistent behavior when both the operating conditions and the measurement realizations differ from the training data.

\subsection{Training Procedure}
\label{sec:training_procedure}
We implement all models in \texttt{Python} using \texttt{PyTorch}~\cite{PyTorch} for automatic differentiation and parameter optimization, and \texttt{torchdiffeq}~\cite{Chen2018} for differentiable ODE integration. We generate all reference trajectories using the mechanistic simulator described in Section~\ref{sec:frp_kinetics}, and the resulting system of differential equations is integrated using the \texttt{LSODA} solver from \texttt{SciPy}~\cite{SciPy}.

We evaluate the three model formulations introduced in Sections~\ref{sec:ml_dynamic_models} and ~\ref{sec:hybrid_NODE}: the discrete-time FNN model, the continuous-time NODE, and the hybrid NODE. In all cases, we optimize the model parameters by minimizing the mean-squared prediction error defined in Equation~\eqref{eq:training}. All trainable parameters are optimized using the AdamW optimizer~\cite{AdamW}.

To promote physically consistent behavior, we include soft penalty terms to discourage nonphysical predictions as they enforce nondecreasing conversion and bounded outputs. We implement these penalties using ReLU-based functions that penalize decreases in conversion or violations of the admissible range $0 \le X \le 1$.

For the discrete-time FNN model, we learn one-step transitions of the form $X_{k+1}$ from $(X_k, u_k, \Delta t_k)$ and generate full trajectories through recursive application. For both regular and irregular sampling scenarios, we train the model for 1500 epochs on the first ten time points. In the unseen-recipe generalization scenario, we train it for 1000 epochs on full trajectories consisting of 25 time points.

For the continuous-time NODE, we parameterize the right-hand side of the ODE as described in Section~\ref{sec:ml_dynamic_models} and train it end-to-end through the ODE solver. For both regular and irregular sampling scenarios, we integrate the system using the \texttt{dopri5} solver and train it for 1500 epochs. In the unseen-recipe generalization scenario, we integrate the system using an explicit \texttt{rk4} scheme and train it for 1000 epochs on full trajectories.

For the hybrid NODE, we retain the mechanistic initiator and monomer mass balances and represent the unresolved radical contribution governing the monomer consumption rate through the learned closure relation described in Section~\ref{sec:hybrid_frp}. We train the hybrid model with the same number of epochs as the data-driven models in each scenario (1500 epochs in the regular and irregular cases, 1000 epochs in the generalization case). We integrate the hybrid system using \texttt{dopri5} in the single-trajectory scenarios and \texttt{rk4} in the generalization scenario.

Across all three evaluation scenarios, we keep the model structures, optimization settings, and training protocols constant within each comparison. Between the first two scenarios, only the measurement sampling pattern is modified.

\section{Results and Discussion}
\label{sec:results}

We evaluate the three model formulations under the scenarios defined in Section~\ref{sec:implementation}. In all scenarios, each trajectory contains 25 time points; training uses the first 10 points (up to $t \approx 0.3$ in normalized time), and evaluation uses the remaining 15 points. Model performance is evaluated by comparing predicted and reference conversion trajectories over the full time horizon.

\subsection{Regular Sampling: Single-Trajectory Reconstruction}
\begin{figure}[H]
    \centering
    \includegraphics[width=0.6\columnwidth]{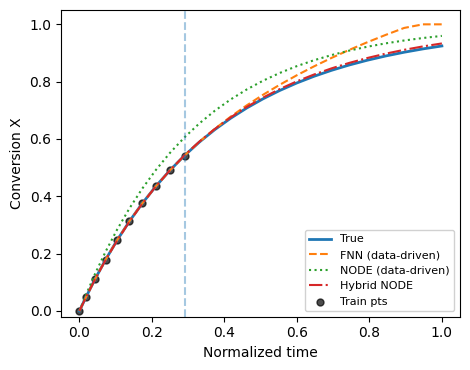}
\caption{Regular sampling, single-trajectory results. Model predictions compared to the mechanistic reference trajectory, highlighting the reconstruction and extrapolation behavior beyond the training region. The vertical dashed line indicates the end of the training region and the beginning of the extrapolation region.}
    \label{fig:regular_traj}
\end{figure}

We first examine a single polymerization trajectory sampled on an equidistant time grid. Figure~\ref{fig:regular_traj} shows the predicted conversion trajectories of the three models together with the mechanistic reference. 

All models follow the reference closely within the training window, where the predictions are directly constrained by the available data. Beyond the training cutoff (indicated by the vertical dashed line), clear differences in extrapolation behavior become apparent.

The discrete-time FNN model formulated according to the discrete-time state-space model introduced in Section~\ref{sec:ml_dynamic_models}, learns a purely data-driven transition mapping between successive time steps. During evaluation, the model is recursively applied to generate the full trajectory. Although this formulation enforces a stepwise temporal structure, it remains entirely unconstrained by mechanistic relations. As shown in Figure~\ref{fig:regular_traj}, the discrete-time FNN model exhibits a pronounced overprediction after the training window and continues to diverge at intermediate and late times. Recursive application of the learned transition mapping amplifies local prediction errors, and the model fails to reproduce the correct curvature and late-stage slowdown of the conversion profile. In this limited-data regime, the discrete-time data-driven formulation does not provide reliable extrapolation.

The fully data-driven NODE produces a smooth trajectory and clearly improves upon the discrete-time FNN model. However, it displays a steeper slope than the mechanistic reference already within the training region. Although it captures the overall trajectory trend more consistently than the discrete-time FNN formulation, it does not accurately reproduce the curvature and late-time slowdown of the mechanistic reference trajectory. The learned continuous-time vector field therefore overestimates the reaction rate within the training window, and this bias accumulates during forward integration. As a result, the model exhibits overprediction at intermediate and late times. While the deviation is less severe than that of the discrete-time FNN model, it remains clearly visible in the trajectory.

In contrast, the hybrid NODE closely follows the reference trajectory over the entire time horizon. After the training cutoff, it maintains a slope consistent with the mechanistic reference and continues to track the conversion trend at late times, indicating improved extrapolation capability compared to both purely data-driven models.
The improved agreement with the mechanistic reference across both the training and extrapolation regions can be attributed to the hybrid formulation introduced in Equation~\eqref{eq:hybrid_general}. Rather than learning the full system dynamics from data, the hybrid NODE only learns the unresolved effective radical contribution governing monomer consumption, while the remaining mass balances and kinetic dependencies are retained from the mechanistic model. As a result, the learning task is significantly simplified. In the limited-data regime considered here, the available measurements are insufficient to reliably identify the full dynamics, as required by the purely data-driven models, but are sufficient to identify the lower-dimensional learned contribution embedded within the mechanistic balances. This leads to improved generalization and physically consistent extrapolation behavior.

In conclusion, the results of Figure~\ref{fig:regular_traj} support the central premise of the proposed hybrid NODE formulation that restricting the learning task to the unresolved kinetic contribution improves the extrapolation capabilities of the model and reduces the amount of data required for accurate prediction even when trained on as few as 10 data points.

\subsection{Irregular Sampling: Single-Trajectory Reconstruction}

We repeated the single-trajectory experiment under irregular sampling while keeping the number of measurements, the training-evaluation split, and all model settings unchanged. Only the measurement schedule was modified such that the time points were no longer regularly spaced. Consequently, the early-time dynamics were observed at irregular intervals, altering how information about the reaction kinetics was represented in the training data.
To isolate the effect of irregular sampling, all model architectures, optimization settings, loss formulations, and training epochs were kept identical to the regular-sampling scenario.

\begin{figure}[H]
    \centering
    \includegraphics[width=0.6\columnwidth]{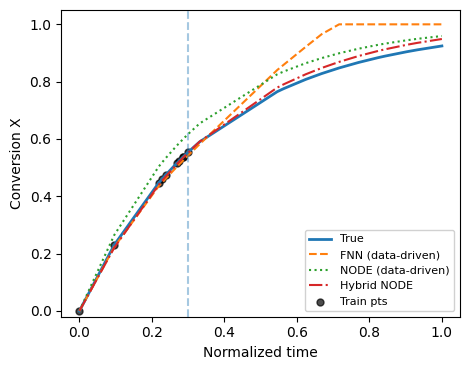}
\caption{Irregular sampling, single-trajectory results. Model predictions are compared to the mechanistic reference trajectory, highlighting the reconstruction and extrapolation behavior beyond the training region. The vertical dashed line indicates the end of the training region and the beginning of the extrapolation region.}
    \label{fig:irregular_traj}
\end{figure}

Figure~\ref{fig:irregular_traj} shows that all models capture the general early-time trend, although deviations are already visible within the training region. In particular, the fully data-driven NODE exhibits a consistently steeper slope than the mechanistic reference, indicating a bias in the learned dynamics. Beyond the training window, the differences in extrapolation behavior become more pronounced.

Under irregular sampling, after the cutoff, the discrete-time FNN systematically overpredicts conversion and predicts complete conversion earlier than the mechanistic reference. Recursive trajectory propagation amplifies local prediction errors, and the varying time increments further increase the sensitivity of the learned transition mapping, resulting in a clear late-time deviation.
The fully data-driven NODE produces smoother trajectories but maintains its slope bias. Nevertheless, it exhibits a slightly steeper slope than the mechanistic reference already within the training window, and this bias is carried into the extrapolation region, resulting in a consistent upward deviation over the full horizon.
Meanwhile, the hybrid NODE remains closest to the reference trajectory over the entire time horizon. While a slight upward deviation is observed at later times, the model preserves the overall curvature despite the irregular distribution of training points, and no systematic drift is observed after the training cutoff.
The improved agreement of the hybrid model with the mechanistic reference is consistent with the formulation introduced in Equation~\eqref{eq:hybrid_general}. Since only the closure term is learned while the mechanistic balances are retained, the model does not need to infer the full system dynamics from irregularly sampled data. 

Under irregular sampling, the early-time dynamics are represented at nonuniform time intervals, which affects how information about the reaction rate is captured. As a result, purely data-driven models learn inconsistent dynamics. The fully data-driven NODE exhibits a biased slope due to the limited and irregularly sampled training data. For the discrete-time FNN, the transition mapping in Equation~\eqref{eq:discrete_nn} depends on the time increment $\Delta t_k$. While this enables the model to handle irregular sampling, it also increases the variability of the learned mapping, as the model must learn transitions across a wider range of time scales. In the sparse-data regime, this leads to increased sensitivity to local errors, which are subsequently amplified after the training cutoff.

The improved agreement of the hybrid model with the mechanistic reference seen in Figure~\ref {fig:irregular_traj} can be attributed to two complementary aspects of the formulation. The continuous-time NODE representation in Equation~\eqref{FRP-hyb} avoids the explicit dependence on the time increments and is therefore less sensitive to irregular sampling. In addition, embedding the neural component within the mechanistic balance restricts the learning task to the unresolved effective radical contribution, which reduces the amount of dynamic information that must be inferred from the limited data. Together, these properties lead to more stable and physically consistent predictions compared to both purely data-driven models.

In summary, irregular sampling increased the difficulty of learning the early-time dynamics, yet the hybrid NODE continued to exhibit the most accurate and physically consistent extrapolation behavior, whereas the purely data-driven models exhibited systematic deviations. These results further support the advantage of embedding mechanistic structure directly within the continuous-time learning formulation.

\subsection{Generalization Across Unseen Recipes Under Noisy Measurements}
In this scenario, we evaluate generalization across unseen recipes under noisy measurements. We generate full conversion trajectories using the mechanistic simulator and perturb the conversion trajectories with additive Gaussian noise ($\sigma = 0.02$). We train all models on the noisy conversion observations and evaluate them against the corresponding noise-free mechanistic trajectories to isolate generalization error from measurement noise.

For each trajectory, we sample the initial monomer concentration $M_0$, initiator concentration $I_0$, chain transfer agent concentration $CTA_0$, and solvent concentration $S_0$ independently from continuous uniform ranges: 
$M_0 \in [3.0, 5.0]$, 
$I_0 \in [0.05, 0.12]$, 
$CTA_0 \in [0.00, 0.06]$, and 
$S_0 \in [3.0, 5.0]$. 
We select training temperatures from the discrete set $\{55, 60\}\,^\circ\mathrm{C}$ and evaluate performance at an unseen temperature of $70\,^\circ\mathrm{C}$. 
The test trajectories, therefore, differ from the training data both in temperature and in the specific combinations of initial conditions, resulting in previously unseen operating conditions. Table~\ref{tab:gen_noise_recipes} lists the representative training and test recipes used for visualization.

\begin{figure}[H]
    \centering

    \begin{subfigure}[t]{0.6\linewidth}
        \centering
        \includegraphics[width=\linewidth]{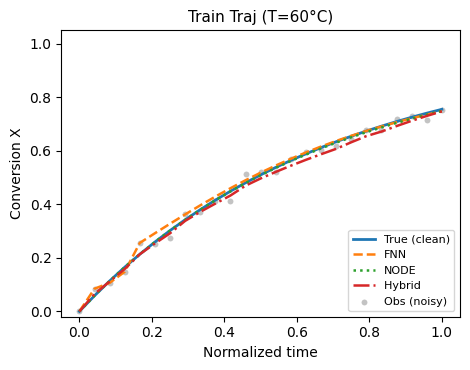}
\caption{Training trajectory (with noisy observations).}
        \label{fig:gen_noise_train_traj}
    \end{subfigure}
    \hfill
    \begin{subfigure}[t]{0.6\linewidth}
        \centering
        \includegraphics[width=\linewidth]{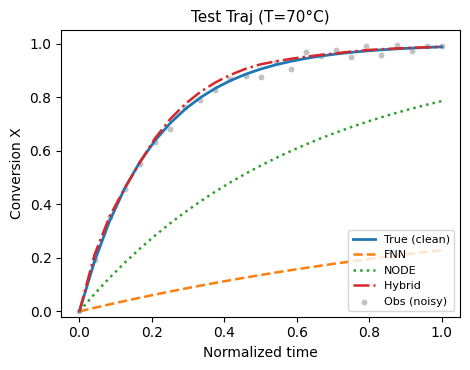}
\caption{Test trajectory at unseen temperature (evaluated against the mechanistic reference).}
        \label{fig:gen_noise_test_traj}
    \end{subfigure}

    \caption{Generalization under noisy data. Model predictions for (a) the training trajectory with noisy observations and (b) the test trajectory at an unseen temperature, highlighting fitting behavior on the noisy training trajectory and generalization to an unseen operating condition.}
    \label{fig:gen_noise_overview}
\end{figure}

We use the same three model formulations introduced previously without modifying the model architectures. We keep the training procedure and the number of training epochs consistent across models, ensuring that performance differences arise from model formulation rather than tuning. Although the models are trained on complete trajectories, the overall dataset remains small, consisting of only two training trajectories with 25 time points each. In Figure~\ref{fig:gen_noise_overview}, we present one representative training trajectory and one representative test trajectory.

\begin{table}[H]
\centering
\caption{Representative training and test recipes for the unseen-operating-condition generalization scenario.}
\label{tab:gen_noise_recipes}
\begin{tabularx}{\columnwidth}{lXX}
\toprule
 & Training trajectory & Test trajectory \\
\midrule
$T$ [$^\circ$C] & 60.0 & 70.0 \\
$M_0$ & 4.3647 & 3.3422 \\
$I_0$ & 0.05377 & 0.11313 \\
$CTA_0$ & 0.01322 & 0.00851 \\
$S_0$ & 3.3687 & 4.2225 \\
\bottomrule
\end{tabularx}
\end{table}
Figure~\ref{fig:gen_noise_overview} summarizes the results for the unseen-condition generalization scenario under noisy training data. 
Subfigure~(a) shows a representative training trajectory at $60^\circ$C, including the noisy observations used for training and the corresponding model predictions. 
Subfigure~(b) presents a representative test trajectory at the unseen temperature of $70^\circ$C, where all models are evaluated against the mechanistic reference. 
Together, these subfigures illustrate how well each model fits the noisy training data and how well the learned dynamics generalize to an unseen operating condition.

In Figure~\ref{fig:gen_noise_train_traj}, all models follow the noisy observations within the training trajectory, which is expected since the models are evaluated on the training dataset. 
The differences become pronounced in the test trajectory in Figure~\ref{fig:gen_noise_test_traj} at the unseen temperature of $70^\circ$C. 
The discrete-time FNN fails to generalize to unseen conditions. Although it fits the noisy training trajectory, it predicts a significantly slower increase in conversion and remains far below the mechanistic reference throughout the batch. This failure to generalize is expected, as the second recipe lies entirely outside the training data, and the discrete-time FNN model, as a purely data-driven transition model, has no mechanism to infer how the dynamics should change under different temperature and initial conditions. Instead, the discrete-time FNN model effectively learns an averaged transition behavior representative of the training conditions and applies it unchanged to the unseen recipe. This limitation leads to severe underestimation of the reaction rate and large prediction errors over the full time horizon.

The continuous-time NODE improves upon the discrete-time FNN formulation but still substantially underestimates the reaction rate at $70\,^\circ\mathrm{C}$. Although the predicted conversion increases more rapidly than for the discrete-time FNN model, the learned dynamics fail to reproduce the accelerated kinetics observed in the mechanistic reference trajectory. Because the NODE must infer the full dynamic vector field from a limited and noisy dataset, the learned dynamics do not extrapolate reliably when both temperature and initial conditions change simultaneously.

In contrast, the hybrid NODE remains in close agreement with the mechanistic reference across the full trajectory. It accurately reproduces the rapid rise in conversion and follows the reference curve without noticeable systematic deviation. This improved agreement arises from the hybrid formulation, in which the mechanistic balances preserve the temperature dependence of the reaction while the neural component learns only the unresolved effective radical contribution governing monomer consumption. As a result, the model generalizes effectively to unseen conditions even when trained on limited and noisy data.

To quantify predictive accuracy, we report the root-mean-square error (RMSE) between the predicted conversion and the mechanistic reference over the full batch horizon. The results in Table~\ref{tab:gen_noise_rmse} clearly support the qualitative observations in Figure~\ref{fig:gen_noise_overview}. The discrete-time FNN exhibits the largest error, reflecting its inability to generalize beyond the training distribution. The fully data-driven NODE reduces the deviation but still shows substantial error, indicating that learning the full system dynamics directly from limited and noisy data remains challenging for reliable extrapolation.

\begin{table}[H]
\centering
\caption{Test performance (RMSE) on the unseen recipe.}
\label{tab:gen_noise_rmse}
\begin{tabularx}{\columnwidth}{lX}
\toprule
Model & RMSE on test trajectory \\
\midrule
FNN & 0.68 \\
NODE & 0.31 \\
Hybrid NODE & 0.013 \\
\bottomrule
\end{tabularx}
\end{table} 

The hybrid NODE achieves a substantially lower error compared to the fully data-driven models, demonstrating accurate prediction of the conversion dynamics under unseen conditions. This result highlights the advantage of the hybrid formulation in addressing one of the central challenges in ML for dynamic systems, namely, learning from limited data. By embedding the neural component within the mechanistic balance equations, the model leverages mechanistic knowledge to constrain the system dynamics while learning only the unresolved effective radical contribution. As a result, the model does not need to infer the full dynamics from data alone but instead complements the available mechanistic structure.

This combination enables the hybrid model to generalize to unseen operating conditions, including different temperatures, while maintaining physically consistent behavior. The results, therefore, show that integrating prior knowledge within the learning framework is not only beneficial for extrapolation but also essential for achieving reliable predictions in data-scarce regimes.

\section{Conclusion and Outlook}
\label{Conclusion and Outlook}
This work demonstrates that a hybrid NODE formulation, in which mechanistic polymerization balances are combined with a learned representation of the unresolved effective radical contribution governing monomer consumption, can accurately predict FRP dynamics under limited-data conditions. Across all evaluation scenarios, the hybrid NODE achieved more accurate and physically consistent extrapolation behavior than the purely data-driven formulations. These results indicate that embedding mechanistic structure directly within the continuous-time learning formulation substantially reduces the amount of dynamic information that must be inferred from data alone. This formulation directly addresses a central challenge in polymerization modeling, namely that key kinetic contributions are only partially known and available data are limited, making it difficult to reliably model the full system dynamics using either purely mechanistic or purely data-driven approaches. In addition, classical mechanistic modeling of FRP typically requires extensive kinetic characterization and parameter estimation, which can become challenging under limited data availability. By combining both, the hybrid formulation leverages existing physical knowledge while using data to learn the remaining unknown contributions.

In the generalization across unseen recipes under a noisy measurements scenario, the hybrid model achieves an RMSE more than an order of magnitude lower than both the discrete-time FNN model and the continuous-time NODE. These results show that learning only the unresolved constitutive contribution governing monomer consumption, rather than the full system dynamics, is sufficient to achieve accurate predictions even with limited, irregular, and noisy data. In contrast to conventional hybrid correction approaches, the proposed formulation does not learn additive corrections to mechanistic trajectories but instead learns the unresolved constitutive contribution required to complete the reduced mechanistic description.

The results also highlight limitations of the purely data-driven formulations considered in this work. Both the discrete-time FNN model and the continuous-time NODE exhibited systematic deviations when evaluated on unseen operating conditions, particularly when extrapolation in both temperature and initial conditions was required. Since these formulations attempt to infer the full system dynamics directly from limited and noisy data, the learned representations remained sensitive to changes in temperature and initial conditions.

These findings align with the growing body of literature on theory-guided and physics-informed ML~\cite{merkelbach2022hybridml,karniadakis2021physics,sorourifar2023physics}. In particular, the superior performance of the hybrid NODE under sparse, noisy, and extrapolative conditions directly supports the observations of Ge et al.~\cite{ge2025machine}, who highlight the limitations of purely data-driven approaches in polymer systems and advocate for hybrid frameworks that embed mechanistic knowledge. The results presented here provide concrete evidence, in the context of FRP, that such structure-consistent hybridization improves both predictive accuracy and generalization. 

At the same time, the present study considers a batch polymerization system and evaluates the methodology using simulator-generated data. The mechanistic simulator, therefore, provides a controlled benchmark for systematically analyzing the influence of sparse, irregular, and noisy data on the different learning formulations. Practical polymerization systems introduce additional challenges, including experimental noise, model mismatch, transport effects, and unobserved disturbances, which were not considered in the present study.
 
Beyond predictive accuracy, the hybrid NODE provides a practical way to extend classical kinetic models. In polymerization systems, several kinetic contributions are often simplified or treated as constants for tractability, despite their known dependence on conversion, chain length, or transport effects. Within the hybrid formulation, such contributions do not need to be specified explicitly. Instead, they can be learned from data through a NODE component embedded within the mechanistic balances, ensuring that the resulting dynamics remain physically consistent while incorporating effects that are difficult to model analytically.

Future work should investigate the application of the proposed hybrid NODE formulation to experimental polymerization datasets, including systems with transport limitations, temperature dynamics, and more complex reaction mechanisms. In addition, extending the framework toward simultaneous prediction of conversion and polymer property distributions, such as molecular-weight distributions, represents an important next step toward practical polymerization process modeling and control. More broadly, the results suggest that hybrid continuous-time learning formulations provide a promising framework for reaction systems in which partial mechanistic knowledge is available but difficult-to-characterize kinetic contributions remain unresolved.

\section*{Acknowledgments}

This project (GA number 101072732) has received funding from the HORIZON-MSCA-2021-DN-01 call of the research and innovation programme of Horizon Europe 2021 under the Marie Skłodowska-Curie actions. The authors acknowledge the support of the Werner Siemens Foundation in the frame of the WSS Research Centre ``catalaix''. \\
We also gratefully acknowledge Prof.~Nicholas Ballard for valuable discussions and guidance on the mechanistic modeling of polymerization processes, as well as Jannik Lüthje for helpful insights related to the systematic formulation of reaction models involving equilibrium and kinetically-limited steps.

\bibliographystyle{elsarticle-num}
\bibliography{references}

\end{document}